\documentclass{article}

    \PassOptionsToPackage{numbers, compress}{natbib}

 \usepackage[position, final]{neurips_2026}

\usepackage[utf8]{inputenc} 
\usepackage[T1]{fontenc}    
\usepackage{hyperref}       
\usepackage{url}            
\usepackage{booktabs}       
\usepackage{amsfonts}       
\usepackage{nicefrac}       
\usepackage{microtype}      
\usepackage{xcolor}         

\usepackage{amsmath}
\usepackage{enumitem}
\usepackage{cleveref}
\DeclareMathOperator*{\argmax}{argmax}
\usepackage[super]{nth}
\Crefname{equation}{Eq.}{Eqs.}

\title{Trustworthy AI Must Account for Interactions}

\author{%
  Jesse C. Cresswell \\
  Layer 6 AI, Toronto, Canada \\
  \texttt{jesse@layer6.ai} \\
}

\begin{document}

\maketitle

\begin{abstract}
  
Trustworthy AI encompasses many aspirational aspects for aligning AI systems with human values, including fairness, privacy, robustness, explainability, and uncertainty quantification. The ultimate goal of Trustworthy AI research is to achieve all aspects simultaneously. However, efforts to enhance one aspect often introduce unintended trade-offs that negatively impact others. In this position paper, we review notable approaches to five aspects and systematically consider every pair, detailing the negative interactions that can arise. For example, applying differential privacy to model training can amplify biases, undermining fairness. Drawing on these findings, we take the position that current research practices of improving one or two aspects in isolation are insufficient. Instead, research on Trustworthy AI must account for interactions between aspects and adopt a holistic view across all relevant axes at once. We offer recommendations for how researchers can work towards integrated trust, and provide guidance for practitioners to manage interactions today.

\end{abstract}

\section{Introduction}
\label{sec:intro}
Artificial intelligence (AI) systems have become widespread for automated decision making and as productivity aids. Industries such as banking increasingly rely on AI models that directly impact customers, while the healthcare sector explores AI-driven advancements in patient care. Increased scrutiny by regulators and concerns around the \emph{trustworthiness} of these systems call for a more measured approach to AI development with considerations beyond raw performance. In response, the field of Trustworthy AI (TAI) has blossomed, with the general goal of aligning AI to human values. Chief among the tenets of TAI are fairness, privacy, robustness, explainability, and uncertainty quantification -- each of which is a noble pursuit, but all of which must be harmonized to promote deep trust. The ultimate goal of TAI as a field is to achieve these aspects \emph{simultaneously}.

These five core concepts are well studied individually. However, as we will discuss, isolated study fails to account for the complex interplay between TAI aspects. Layering multiple methodologies designed for individual aspects tends to produce negative interactions, ultimately undermining trust rather than reinforcing it. Many such negative interactions have been documented, but their prevalence and severity may not yet be fully realized due to the diverse and siloed nature of research on TAI. By collating them in one place, we aim to bring these interactions to light and highlight that a more holistic approach to TAI is needed. \textbf{We take the position that TAI must account for interactions between aspects to achieve its stated goals of aligning AI with human values.}

\section{Trustworthy AI Aspects}
\label{sec:apsects}
Throughout the discussions below we will typically consider a classification model $F_\theta: \mathcal{X}\to \mathcal{Y}$ parameterized by $\theta$, which maps a feature space $\mathcal{X}$ to a discrete set of labels $\mathcal{Y}$. We use capital letters (e.g. $X$ and $Y$) to denote random variables, while lower case (e.g. $x\in\mathcal{X}$ and $y\in\mathcal{Y}$) indicates data instances. $F_\theta$ is trained to minimize a loss function $\mathcal{L}$ over its training set $\mathcal{D}_\text{train}$. Softmax outputs are denoted as $f_\theta: \mathcal{X}\to [0,1]^m$, such that $F_\theta(x) = \argmax_{i \in \mathcal{Y}} f_\theta(x)_i$. 

We first overview five TAI aspects whose interactions we consider in \Cref{sec:pairwise}. These reviews are intentionally \emph{selective}, not comprehensive, and focus on a limited set of topics to highlight that negative interactions are commonplace, not arbitrarily chosen from a wide-ranging body of literature.
\subsection{Fairness}\label{sec:fairness}
Fairness is a foundational pillar in the development of TAI, ensuring systems treat diverse populations equally or equitably. Since fairness is a nuanced and highly contextual topic, it cannot be boiled down into a single set of guidelines to follow in all cases. Instead, TAI researchers and practitioners must consider the appropriate fairness definitions and methodologies to use in each circumstance. 

We review the case where the data is partitioned into $n_\text{g}$ groups based on an attribute $a\in \mathcal{A}=\{1, ..., n_\text{g}\}$ (e.g. age bins), and consider the viewpoint of substantive fairness which aligns with the concept of equity. It encourages treating individuals differently to achieve comparable outcomes~\cite{mittelstadt2024unfairness}. If individuals do not receive the same beneficial outcomes from a model, there is \emph{Disparate Impact}, which is a commonly used concept in legal and regulatory frameworks \cite{occ_fair_lending}. It can be measured in terms of the target outcome of the model, for instance through \emph{accuracy disparity}, where $\mathcal{D}_a$ denotes the subset of $\mathcal{D}$ belonging to group $a$:
\begin{equation}\label{eq:disparate_impact}
    \Delta_\text{acc} = \max_{a, b \in \mathcal{A}}[\mathrm{acc}(F_\theta, \mathcal{D}_a) - \mathrm{acc}(F_\theta, \mathcal{D}_b)].
\end{equation}
Minimizing $\Delta_\text{acc}$ is one example of a fairness goal, and can be pursued at several stages including pre-processing (e.g. balancing data across groups before training), in-processing (e.g. adding fairness regularization terms to the loss $\mathcal{L}$), or post-processing (e.g. using model scores differently across groups when making decisions). These \emph{fairness interventions} are examples of intentionally treating groups differently so that outcomes will be more similar.
\subsection{Privacy}\label{sec:privacy}
Privacy has become a crucial area of research in TAI as systems increasingly rely on sensitive data for training \cite{liu2021privacy}. The use of personal information, such as social media activity, has led to growing concerns about privacy breaches, unauthorized data access, and the risk of re-identification \cite{carey2023reidentification}.

In the context of ML, privacy concerns are usually demonstrated adversarially, where an attack is employed to extract as much private information as possible from the model itself, or its outputs. The standard example is a \emph{membership inference attack} (MIA) \cite{shokri2017, ye2022mia} in which the adversary tries to determine if a test datapoint $x_\text{test}$ was included in $\mathcal{D}_\text{train}$. MIAs help to demonstrate when a system can fail to preserve privacy, but an unsuccessful MIA does not indicate the system is safe; there could always exist a stronger attack that would succeed. Hence, privacy researchers rely on future-proof frameworks that provide statistical guarantees on privacy protection.

\emph{Differential privacy} (DP) \citep{dwork2006} is the primary framework for quantifying how much private information could be exposed by an ML model. Formally, let $M$ be a probabilistic function acting on datasets $\mathcal{D}$. We say that $M$ is $(\epsilon, \delta)$-\emph{differentially private} if for all subsets of possible outputs $S\subseteq \text{Range}(M)$, and for all pairs of datasets $\mathcal{D}$ and $\mathcal{D}'$ that differ by the addition or removal of one element,
\begin{equation}\label{eq:dp}
    \text{Pr}[M(\mathcal{D})\in S] \leq \exp(\epsilon)\, \text{Pr}[M(\mathcal{D}')\in S] + \delta.
\end{equation}
This inequality guarantees that the function $M$ cannot strongly depend on any one datapoint, and hence the amount of information that can be extracted about any datapoint is bounded. Strong DP guarantees (i.e. $\epsilon$ and $\delta$ both close to 0) have been empirically shown to be effective defenses against MIAs and other privacy attacks \cite{rahman2018membership, ye2022privacy}. Importantly, no amount of post-processing on the outputs of $M$ can weaken its guarantee. DP is typically achieved in ML through DPSGD \citep{Abadi_2016}, an optimization method that satisfies \Cref{eq:dp}. It first computes per-sample gradients and clips them, then aggregates them before adding noise. For samples $x_i, y_i$ in a batch $B$, the DPSGD gradient update is
\begin{equation}\label{eq:dpsgd}
\theta_{t+1}=\theta_{t} - \lambda \bigg[\frac{1}{\vert B \vert}\sum_{i\in B} \textrm{clip}_C\left(\nabla_{\theta}\mathcal{L}(\theta_t; x_i, y_i)\right) + \frac{\sigma C}{\vert B\vert}\xi\bigg], \ \  \mathrm{clip}_C(g) := g \cdot \min\left(1,\; \frac{C}{\|g\|_2}\right),
\end{equation}
where $\lambda$ is the learning rate, $C$ is the clipping bound, $\sigma$ is the noise level, and $\xi \sim \mathcal{N}(0, \textbf{I})$ is Gaussian noise. As training with DPSGD progresses, more \emph{privacy budget} is consumed ($\epsilon$ and $\delta$ increase), which is usually accounted numerically \cite{mironov2019renyi, yousefpour2021opacus}.

\subsection{Robustness}\label{sec:robustness}
Robustness broadly refers to the ability of a model to maintain its performance and reliability under a variety of conditions. Since stable performance is desired even in unforeseen situations, researchers commonly test robustness adversarially. An attacker actively tries to produce unintended behaviour by perturbing inputs, often in ways that are imperceptible to humans \cite{biggio2013evasion, szegedy2014intriguing}. \emph{Adversarial examples} can be created through optimization by maximizing the model's loss on the correct answer:
\begin{equation}
    x^\dagger(\theta, x_i, y_i) = \argmax_{x\in \mathcal{B}(x_i, \varepsilon)} \mathcal{L}(\theta; x, y_i),
\end{equation}
where $x^\dagger$ is constrained to be ``close'' to $x_i$, e.g. within a ball $\mathcal{B}(x_i, \varepsilon)$ of radius $\varepsilon$ around $x_i$. Such attacks can be defended against by exposing the model to adversarial inputs during training \cite{goodfellow2015explaining}:
\begin{equation}\label{eq:adversarial_training}
\theta_{t+1}=\theta_{t} - \lambda \frac{1}{\vert B \vert}\sum_{i\in B} \nabla_{\theta} \mathcal{L}(\theta_t; x^\dagger(\theta_t, x_i, y_i), y_i).
\end{equation}
\subsection{Explainability}\label{sec:explainability}
Explainability enables researchers and practitioners to understand, validate, and trust decisions made by complex models, and gives the ability to audit those decisions retroactively. When humans manually accept or reject a model's prediction, explanations help them understand the reasoning behind the decision and compare it to their own expertise. Some ML models are inherently more interpretable, such as shallow decision trees and linear models, but deep neural networks are not. Methods to explain complex models often focus on \emph{local explanations} which shed light on individual predictions \cite{linardatos2021ex}. We will focus on model-agnostic, feature-based explanations due to their applicability across algorithms \citep{islam2021explainable}. These methods interpret behavior by analyzing the importance of input features for a given prediction, regardless of the underlying model architecture, by providing some form of \emph{feature importance} $E_\theta(x)$, a quantification of how important each element of $x$ is for predicting $F_\theta(x)$.

We recount one method as an example, Local Interpretable Model-agnostic Explanations (LIME) \cite{ribeiro2016lime}. LIME aims to provide local explanations that preserve local \emph{fidelity}---that the explanations correspond to the model's actual behaviour in the vicinity of $x$. LIME's explanations take the form of a sparse linear model that locally approximates $F_\theta$ at $x$, but which is inherently more interpretable than $F_\theta(x)$. The weights of the linear model are returned as $E_\theta(x)$ and communicate feature importance.
\subsection{Uncertainty Quantification}\label{sec:uq}
Typical ML models output a point prediction and are not designed to quantify confidence in those predictions. We note that softmax outputs $f_\theta(x)$ are unreliable because of miscalibration \cite{guo2017calibration, minderer2021revisiting}. ML models that quantify their uncertainty are more trustworthy, as the user can judge when to ignore the model in favor of alternatives \cite{soize2017uncertainty}. We focus on one increasingly popular method, \emph{conformal prediction} (CP) \cite{vovk2005algorithmic}. The idea of CP is to output sets of predictions (e.g. several class labels) where larger sets indicate greater model uncertainty. CP takes a heuristic notion of uncertainty, like $f_\theta(x)$, and calibrates it using a held-out dataset $\mathcal{D}_\text{cal}$. CP defines a \emph{conformal score} function $s:\mathcal{X} \times \mathcal{Y}\to \mathbb{R}$, where larger values indicate worse agreement between $f_\theta(x)$ and $y$. After computing $s$ on the $n_\text{cal}$ calibration datapoints, one finds the $\tfrac{\lceil{(n_\text{cal}+1)(1-\alpha)}\rceil}{n_\text{cal}}$ quantile $q$ of the scores, using a free parameter $\alpha\in (0,1)$. For a new datapoint $x_\text{test}$, prediction sets $\mathcal{C}_{q}$ are generated by including all output values for which the conformal score is below the threshold $q$,
\begin{equation}\label{eq:prediction-set}
    \mathcal{C}_{q}(x_\text{test})  = \{y\in \mathcal{Y} \mid s(x_\text{test}, y) < q \}.
\end{equation}
Notably, CP provides a \emph{coverage} guarantee over the true label $y_\text{test}$, where $\alpha$ is the error rate,
\begin{equation}\label{eq:coverage-guarantee}
     \mathbb{P}[y_\text{test} \in \mathcal{C}_q(x_\text{test})] \geq 1 - \alpha,
\end{equation}
as long as $x_\text{test}$ is exchangeable with the calibration data drawn from $\mathbb{P}$. For equal coverage levels smaller average set sizes $\mathbb{E}\vert \mathcal{C}_{q}\vert$ are more useful, and indicate more confident predictions \cite{romano2020classification, angelopoulos2021raps, huang2024saps, cresswell2024icml}.
\subsection{Other Aspects}\label{sec:otheraspects}
Our discussion focuses on the five aspects recounted above. There are, however, \emph{many} additional aspects one may strive to achieve when building TAI which we mention only in passing. 
\textbf{Safety} may mean ensuring that AI systems cannot cause harm to users, subjects, society, or the environment \cite{amodei2016concrete, hendrycks2024safety}. 
\textbf{Alignment} focuses on ensuring that the objectives and behaviours of an AI agent are consistent with human values and societal goals \cite{russell2019human, gabriel2020artificial, sorensen2024alignment}. 
\textbf{Diversity} entails incorporating diverse perspectives, identities, and contexts in the design, development, and deployment of AI systems \cite{buolamwini2018gender, fazelpour2022diversity}. 
\textbf{Reproducibility} allows outputs from AI to be replicated on demand which is important for the advancement of AI science \cite{pineau2021reproducibility}, but just as importantly allows real-world deployments of AI to be audited.
\textbf{Accountability} clearly assigns responsibility for AI decisions and their outcomes, providing legal recourse when AI systems cause harm \cite{cooper2022accountability}. 
\textbf{Human agency} means ensuring that AI systems empower individuals to make informed decisions, rather than overriding human autonomy \cite{fanni2023enhancing}. 
Other types of models introduce their own desirable aspects. As an example type of generative model, large language models (LLMs) should not hallucination, abstain when uncertain, output only safe content, and be robust to jailbreak attempts. 
There are still more desirable qualities, like accessibility or adaptability, that we should strive for, but are omitted here for brevity.

\section{Negative Interactions Between Trustworthy AI Aspects}
\label{sec:pairwise}
\vspace{-6pt} 
As is evident from even the brief overview in \Cref{sec:apsects}, many technical solutions have been proposed to improve individual TAI aspects. However, these solutions often unintentionally harm other aspects, inhibiting progress towards the overall goal of achieving all aspects simultaneously. To demonstrate that negative interactions are commonplace, not the exception, we exhaustively consider every pair of our five aspects: Fairness (F), Privacy (P), Robustness (R), Explainability (E), and Uncertainty Quantification (UQ). For each pair we give examples of negative implications on one aspect from the application of the other, and cover both directions. For two TAI aspects A and B, we use the shorthand A~$\rightharpoonup$~B to indicate that applying a concept or method from A has a negative impact on B.

While there are also examples of positive interactions, we take a risk management viewpoint and focus on negative interactions to demonstrate the potential harms of not approaching TAI holistically. Also, when there are alternative or competing notions within a single aspect that should be achieved simultaneously, internal interactions can also occur. For example, in fairness it is provably impossible to simultaneously satisfy several common definitions of group fairness~\citep{chouldechova2017fair}. Below, we limit the discussion to inter-aspect interactions, though similar arguments apply to intra-aspect interactions.
\subsection{Fairness and Privacy}\label{sec:FP}
\textbf{F $\rightharpoonup$ P:} At the most basic level, evaluating or correcting the fairness of an ML model with respect to some group usually necessitates collecting information on the group identifier $\mathcal{A}$. These identifiers, like age, gender, or race, are often sensitive personal information -- exactly the type of information that should be afforded privacy. Collecting, storing, and using this information for fairness purposes exposes individuals to greater risks of conventional data leaks or hacks.

Beyond conventional privacy leaks, \Cref{sec:privacy} discussed how trained models can leak private information through MIAs which exploit the differences in model behaviour between populations. Fairness interventions during training can reduce the differences between populations, and hence better protect against standard MIAs \cite{tonni2020data}. However, such techniques actually increase vulnerability to specialized MIAs which are also group-aware \cite{tian2024fairprivate}. 
Generally, fairness-aware training methods, such as adding a fairness term or constraint to the loss, force a model to pay more attention to underrepresented groups, but this leads to stronger memorization of those datapoints, which are then easier to extract information about in MIAs \cite{chang2021privatefair}.

\looseness=-1\textbf{P $\rightharpoonup$ F:} Some individuals or groups in the data can be more vulnerable to privacy attacks \cite{long2020privatefair}. When vulnerability is unequal, applying privacy-enhancing techniques can improve the privacy of some groups more than others, an example of Disparate Impact \cite{kulynych2022disparate}. Protecting a vulnerable group by removing it from training is counterproductive, as vulnerability merely shifts to a different group \cite{carlini2022onion}.

\looseness=-1 While DPSGD is the \emph{de facto} standard method for achieving privacy guarantees on ML models, it is well-known to cause Disparate Impact by increasing accuracy disparity (\Cref{eq:disparate_impact}) \cite{bagdasaryan2019}. Suppose some group $a\in\mathcal{A}$ in the data is underrepresented, or more difficult to correctly predict on. In ordinary SGD, datapoints from this group would have higher loss, and hence larger gradients, which would increase their relative influence on the optimization. In DPSGD (\Cref{eq:dpsgd}), large gradients with $\Vert g_i \Vert > C$ are clipped, making them relatively less influential on the optimization process. This uneven clipping introduces bias into the gradients which is the primary source of Disparate Impact \cite{tran2021a, esipova2023disparate}.
\subsection{Fairness and Robustness}\label{sec:FR}
\textbf{F $\rightharpoonup$ R:} When groups in the dataset are underrepresented, fairness interventions to reduce model bias (\Cref{sec:fairness}) can increase the relative influence of those groups. Unfortunately, this increased influence can make the very same groups more susceptible to adversarial attacks \cite{chang2020adversarial, xu2021robustfair}. \citet{tran2024fairrobust} show that fairness interventions can reduce the average distance from training samples to the decision boundary, which makes them more vulnerable to adversarial examples \cite{madry2018towards}.

\textbf{R $\rightharpoonup$ F:} The main method to improve adversarial robustness, namely adversarial training, adds perturbed versions $x^\dagger$ of inputs $x$ to training batches (\Cref{eq:adversarial_training}). The side-effects of adversarial training include decreased overall accuracy on unperturbed samples, but more importantly for fairness,  larger disparities in class-wise performance  \cite{nanda2021robustfair, xu2021robustfair, benz2021robustfair}. This robustness bias has been attributed to properties of the data distribution like feature distributions across groups \cite{benz2021robustfair}, differences in the intrinsic difficulty of classes \cite{xu2021robustfair}, and biased representations learned during pre-training \cite{nanda2021robustfair}.

\subsection{Fairness and Explainability}\label{sec:fe}

\textbf{E $\rightharpoonup$ F:} Explanations introduce another potential source of bias in modeling. Even if a model’s predictions are considered fair, the fidelity of explanations may be inconsistent across groups -- that is, for some groups the features identified as important in the explanation may not truly reflect the features driving the model’s predictions. Fidelity disparity can lead to the model's predictions being trusted more for some groups than others, such that the benefits of the model are not evenly experienced across groups \cite{balagopalan2022explainfair}. \citet{dai2022fairexplain} found that \emph{post hoc} explanation methods used on neural networks quite commonly have disparate fidelity across groups.

Alternatively, explanations may hide biases in an unfair model. For instance, explanations may fail to represent that a model is relying on sensitive attributes, covering up active discrimination \cite{lakkaraju2020explain, slack2020robustexplain}.
\subsection{Fairness and Uncertainty Quantification}\label{sec:fuq}
\textbf{Background:} In conformal prediction the coverage guarantee in \Cref{eq:coverage-guarantee} holds marginally over the entire distribution $\mathbb{P}$. Hence, some groups within the distribution may have lower coverage than others, leading to tension between fairness and UQ. A stronger guarantee is \emph{group-wise conditional coverage} with respect to pre-defined groups $\mathcal{A}$,
\begin{equation}\label{eq:conditional-coverage}
     \mathbb{P}[y \in \mathcal{C}(x) \mid A=a] \geq 1 - \alpha, \quad \forall \ a\in \mathcal{A}.
\end{equation}
Group-wise conditional coverage can easily be obtained by partitioning $\mathcal{D}_{\text{cal}}$ by groups, and performing CP on each $\mathcal{D}_{a}$, giving distinct thresholds $q_a$ \cite{vovk2003mondrian}.

\textbf{F $\rightharpoonup$ UQ:} The idea of providing equal levels of coverage across groups for the sake of fairness was discussed by \citet{romano2020with}, who argued that \emph{Equalized Coverage} should be the standard of fairness for CP. Using notation similar to \Cref{eq:disparate_impact}, we can express Equalized Coverage as
\begin{equation}\label{eq:delta_cov}
    \Delta_\text{Cov}=\max_{a, b\in \mathcal{A}}\left( \mathbb{P}[y \in \mathcal{C}(x) \mid A=a] - \mathbb{P}[y \in \mathcal{C}(x) \mid A=b]\right)\approx 0.
\end{equation}
Marginal coverage gives no guarantee that \Cref{eq:delta_cov} will hold, but group-wise conditional coverage does. However, Equalized Coverage negatively impacts the usefulness of prediction sets for uncertainty quantification by increasing their average size, meaning the model expresses a greater level of uncertainty than it would using marginal CP \cite{romano2020classification, gibbs2023conformal, ding2024class}. Additionally, partitioning $\mathcal{D}_{\text{cal}}$ means each individual calibration is done with fewer datapoints $n_\text{cal}$. This increases variance, and the probability that the desired coverage level $1-\alpha$ is breached in practice \cite{angelopoulos2022gentle}.

\textbf{UQ $\rightharpoonup$ F:} 
Prediction sets can be used to assist human decision makers \cite{straitouri2024designing, cresswell2024icml}. The usefulness of prediction sets is correlated to set size---humans have higher accuracy on tasks when given smaller prediction sets \cite{cresswell2024icml}. Average set sizes $\mathbb{E}\vert \mathcal{C}_{q}\vert$ typically vary across groups when the underlying model $f_\theta$ has some accuracy disparity $\Delta_\text{acc} > 0$ (\Cref{eq:disparate_impact}). As a result, human accuracy will improve more for groups which have smaller sets on average, causing Disparate Impact \cite{cresswell2024conformal}. Equalized Coverage makes this unfairness worse. If a group in the data is under-covered using marginal CP, equalizing its coverage requires increasing set sizes, harming downstream accuracy even more.
\subsection{Privacy and Robustness}\label{sec:pr}
\textbf{P $\rightharpoonup$ R:} Models trained with DPSGD (\Cref{eq:dpsgd}) tend to be less adversarially robust than the same models trained without DP guarantees. The clipping and noise addition steps in DPSGD slow the convergence of models~\cite{tramer2021differentially} giving decision boundaries that are less smooth~\cite{hayes2022learning} which has a strong impact on adversarial robustness~\cite{fawzi2018empirical}. Empirical tests confirm this intuition; \citet{boenisch2021gradient} show examples where DPSGD caused MIA success rates to increase to 95\% from 19\% in an SGD baseline. 

\textbf{R $\rightharpoonup$ P:} Adversarial training (\Cref{eq:adversarial_training}), designed to improve adversarial robustness, can increase the influence of individual datapoints on the model. This in turn makes the model more susceptible to MIAs \cite{yeom2020overfitting}. \citet{song2019privaterobust} tested six common adversarial defence methods and found all six increased the success rates of MIAs compared to the same model trained without any specific defence.

Incorporating DP alongside adversarial defences to protect against MIAs is also non-trivial, as the methodologies conflict in practice. Adversarial training creates several augmentations $x^\dagger$ of datapoints $x$, and backpropagates gradients over them in a batch. By comparison, DPSGD computes per-sample gradients, which is on its own computationally inefficient \cite{yousefpour2021opacus}. Incorporating augmented datapoints drastically increases the time and memory costs of training, and requires careful accounting of how much privacy budget is consumed by the use of augmented versions of $x$ \cite{wu2024augment}.
\subsection{Privacy and Explainability}\label{sec:pe}
\textbf{P $\rightharpoonup$ E:} DPSGD is designed to obscure the details of any single element of $\mathcal{D}_\text{train}$, but its addition of noise to gradient updates can degrade the fidelity of \emph{post hoc} explanations by clouding the true relationships between input and output variables \cite{patel2022dpxai}. \citet{saifullah2024privacy} found severe deterioration of explanation fidelity across many model architectures and data domains when DPSGD was used.

Applying DPSGD during training protects elements of $\mathcal{D}_\text{train}$, but not inference data $x_\text{test}$ whose predictions need to be explained. DP can be applied to the explanation mechanism to protect $x_\text{test}$ \cite{patel2022dpxai}, but since DP requires randomization, explanations will necessarily differ each time they are generated for the same $x_\text{test}$. Unstable and potentially inconsistent explanations undermine the premise that we can understand the true reasons behind model predictions.

\textbf{E $\rightharpoonup$ P:} Local explanations $E_\theta(x)$ pose an additional avenue for private information about $x$ or $\mathcal{D}_\text{train}$ to leak from the model. Explanations are \emph{designed} to reveal details about how specific inputs influence model predictions, so it is unsurprising that they can be exploited to make MIAs more effective \cite{shokri2021privacyexplain}. The explanations generated by LIME (\Cref{sec:explainability}) consist of a simple model that locally approximates $F_\theta(x)$ around $x$. The behaviour of the local model will vary depending on whether $x$ was included in $\mathcal{D}_\text{train}$, and attackers can exploit these differences in their MIAs \cite{quan2022amplification, huang2024explaining}.
\subsection{Privacy and Uncertainty Quantification}\label{sec:puq}
\textbf{P $\rightharpoonup$ UQ:} Conformal Prediction could be applied to any model trained with DPSGD without affecting the privacy of $\mathcal{D}_\text{train}$, because CP is merely post-processing on the privatized model $f_\theta(x)$. However, CP requires a calibration set $\mathcal{D}_\text{cal}$ which would not inherit any DP guarantee. To mitigate the risk to $\mathcal{D}_\text{cal}$, one can generate \emph{private prediction sets} via a DP quantile routine \cite{Angelopoulos2022Private}, such that prediction sets $\mathcal{C}_q(x)$ would satisfy \Cref{eq:dp} for $\mathcal{D}_\text{cal}$ and $\mathcal{D}_\text{cal}'$ differing by one element. However, the noise added for DP degrades the empirical coverage of $\mathcal{C}_q(x)$, requiring larger prediction sets to retain the same coverage $1-\alpha$, hence overestimating the true model uncertainty.

Even without protecting the privacy of $\mathcal{D}_\text{cal}$, the quality of UQ with a differentially private model will suffer compared to a non-private model. The per-example clipping used in DPSGD causes miscalibration \cite{bu2023on, zhang2022closer}, which affects the utility of CP by increasing the size of prediction sets \cite{xi2024does}.

\textbf{UQ $\rightharpoonup$ P:} Uncertainty quantification techniques by design provide additional information to supplement the model's prediction, which broadens the attack surface. As proof of concept, \citet{zhu2024uncertainty} developed and tested MIAs targeting prediction sets, showing empirically that an attacker with prediction sets has a higher success rate.
\subsection{Robustness and Explainability}\label{sec:re}
\textbf{R $\rightharpoonup$ E:} Adversarial training fundamentally alters the representations that are learned by $F_\theta(x)$ \cite{tsipras2018robustness, zhang2019interp}. In image classification, features learned with adversarial training are often more interpretable to humans \cite{ilyas2019bugsnotfeatures}, but other data modalities do not share the same alignment between robust features and human-perceptible patterns \cite{jia2017text, Carlini2018audio}. In such domains, adversarial training can lead to unexplainable behaviours and reduced explanation fidelity \cite{ZHOU2025109681}.

\textbf{E $\rightharpoonup$ R:} \emph{Post hoc} explanations are susceptible to adversarial perturbations which do not change the model's prediction $F_\theta(x)$, but greatly change the explanation $E_\theta(x)$ \cite{ghorbani2019explainrobust}. Users may expect there to be a single explanation for any given prediction, and hence non-robust explanations cast doubt on the veracity of all explanations. Alternatively, adversarial examples can be used to generate explanations from methods like LIME (\Cref{sec:explainability}) which are not faithful to the model's actual behaviour \cite{slack2020robustexplain}.
\subsection{Robustness and Uncertainty Quantification}\label{sec:ruq}
\textbf{R $\rightharpoonup$ UQ:} 
Conformal prediction sets may fail to be robust if the underlying model is non-robust. Standard CP methods are based on softmax scores $f_\theta(x)$ \cite{romano2020classification}, so if the elements of $f_\theta(x_\text{test}^\dagger)$ vary wildly, $\mathcal{C}_q(x_{\text{test}}^\dagger)$ will as well. Hence, adversarial training on the underlying model may appear to be a natural defence for CP. However, \citet{liu2024conformaladversarial} demonstrated that adversarial training increases the overall uncertainty of models, leading to larger set sizes even for clean datapoints $x_\text{test}$.

\textbf{UQ $\rightharpoonup$ R:} CP techniques are highly susceptible to adversarial attacks because they introduce additional assumptions which can easily be violated by an attacker. The coverage guarantee (\Cref{eq:coverage-guarantee}) relies on exchangeability of $x_\text{test}$ with $\mathcal{D}_{\text{cal}}$, but adversarial perturbations can imperceptibly force $x_\text{test}$ out-of-distribution \cite{gendler2022adversarially}. Prediction sets under attack will grossly overestimate the certainty of predictions or fail to cover the true label. Alternatively, $x_\text{test}$ can be perturbed such that coverage is maintained, but prediction set sizes are greatly increased, which reduces the utility of those sets \cite{ghosh2023robustconformal}.
\subsection{Explainability and Uncertainty Quantification}\label{sec:euq}
\textbf{E $\rightharpoonup$ UQ:} When generating explanations of a model's predictions, it may be necessary to quantify the uncertainty in the explanations themselves, as high uncertainty about the validity of explanations can erode trust \cite{Kindermans2019, bykov2020much, ahn2023uncertainty, lofstrom2024explaincalibration}. For example, \citet{slack2021explainuq} highlighted that the feature importances $E_\theta(x)$ generated by LIME strongly depend on the random noise introduced when constructing the sparse linear model around $x$, and on the number of perturbed samples used. These factors can lead to significant variations in the rank order of important features, indicating a considerable degree of uncertainty in LIME's explanations that is often overlooked.

Moreover, methods like LIME operate on point predictions and do not account for the model's uncertainty in $F_\theta(x)$. Providing explanations when the model itself is highly uncertain may communicate a false sense of confidence, potentially misleading users to rely on doubtful predictions \cite{gosiewska2019not, zhang2019should, slack2020robustexplain}.

\textbf{UQ $\rightharpoonup$ E:} To understand the limitations of a model, one should be able to explain why it is more uncertain on some inputs than others \cite{antoran2021getting}. For CP this means extending explanations to prediction sets $\mathcal{C}_q(x)$. However, the task of explaining why the model has predicted the entire set is inherently more difficult than explaining the top prediction $F_\theta(x)$, especially when elements of the set may be contradictory or incompatible \cite{yapicioglu2024euq}.

\subsection{Other Aspects}
Finally, we note that although we have focused on examples of negative interactions in classification models, other types of models exhibit similar concerns. Following the LLM example from \Cref{sec:otheraspects}, fine-tuning an LLM for alignment on a specific topic can subvert pre-training guardrails, compromising safety and jailbreak resilience \citep{ICLR2024_83b7da3e}. Reducing hallucinations by fine-tuning on benign data can similarly harm abstention behaviour \citep{mahmoud2026unintended}. We do not intend this review to be exhaustive, but to illustrate the negative interactions among the most widely recognized and studied aspects of TAI.

\section{Position: Trustworthy AI Must Account for Interactions}
\label{sec:position}
The overall goal of Trustworthy AI research is to enable not one or two aspects of trust, but \emph{many} simultaneously. Current research in TAI very commonly follows the same formula: one or two TAI aspects are selected and genuine issues with AI models are used to motivate improving these aspects. Then, technical solutions are developed and evaluated to show improvement on the selected aspects; possible interactions with aspects outside the ones selected are rarely considered.

The most straightforward attempt to achieve the overall goal of trust would be to overlay several technical solutions. However, negative interactions compromise this attempt since solutions implemented later undermine earlier ones. This would not be so much of a problem if interactions were rare, but the examples from \Cref{sec:pairwise} demonstrate that negative interactions between TAI aspects are pervasive---on a scale that has not previously been recognized. Here, the selectivity of our survey is important; the fact we can find negative interactions between all pairs in both directions while limiting ourselves to a single technical solution per aspect strongly suggests that negative interactions are commonplace and pervasive. To put it another way, if negative interactions are pervasive, one should be able to find them in all combinations for a given technical solution, and we do.

Based on these observations we take the position that combining solutions to individual TAI aspects will not resolve the trust and alignment problems facing AI. Trustworthiness is not achieved by overlaying isolated technical solutions, but emerges from integrating TAI aspects within a holistic framework that accounts for interactions. TAI should consider all relevant aspects simultaneously, not in a sequential or siloed manner. This will rely on interdisciplinary expertise from ethicists, legal experts, and computer scientists, to bring together knowledge from disparate fields. TAI must be context-aware and adapted to its deployment domain, taking account of specific requirements -- like the primacy of patient safety in healthcare. Trust in AI systems will be achieved not solely through technical solutions, but by aligning AI with societal needs, and recognizing real-world constraints.

\subsection{Call to Action: Steps to Advance TAI for Researchers}
Given that current research practices are insufficient to achieve holistic trust, we propose several concrete steps researchers can take that will advance TAI toward its overall goal:
\begin{enumerate}[leftmargin=*, itemsep=2pt, topsep=2pt]
    \item \textbf{Multi-Aspect Benchmarks}: Researchers must create standardized repositories of datasets and metrics that facilitate multi-aspect evaluation of TAI solutions. This will enable efficient ablations across aspects to identify potential negative interactions, map out the multi-dimensional Pareto frontier of trust, and improve transparency on whether new techniques strictly improve the TAI landscape, or merely shift the burden to another trust axis. Datasets curated for this type of benchmark should expose intersectional tensions, for example having underrepresented groups, a variety of private information, and ambiguous data where uncertainty quantification is necessary. Researchers should set standards for reporting impact along multiple axes to prevent newly proposed solutions from deliberately not reporting unsatisfactory results about any given aspect.

    \item \textbf{Holistic TAI Venues}: Existing systems of ML research reward relatively short papers via strict page limits, and state-of-the-art performance on benchmarks. Holistic TAI research in contrast will require expansive consideration of many topics, and the management of trade-offs in performance. Hence, we encourage the establishment of workshops and publication venues with policies that facilitate holistic TAI research. This can include extended page limits, Calls for Papers that explicitly target the intersection of trust aspects, and expectations to benchmark across many aspects. An important deviation from existing practices will be to redefine criteria for success.     Qualities that could be encouraged include breadth, thoroughness of documenting side effects, consideration of the entire ML model lifecycle (data → training → deployment → monitoring), and inclusion of interdisciplinary experts such as ethicists, legal scholars, and sociologists.

    \item \textbf{Prioritize Humanity}: Trustworthy AI is fundamentally about sociotechnical integration, not purely technical advancement of capabilities. It depends on human oversight, governance, and context. Researchers should study how to incorporate human oversight mechanisms with technical research contributions, and how humans can play a bridging role across aspects, acting as a backstop when aspects are in tension. Also, researchers should deliberately consider how TAI research is communicated to those who may use it. An infamous example of insufficient communication comes from DP guarantees stated in terms of $\epsilon$ and $\delta$ (\Cref{eq:dp}). The high technical bar for understanding the consequences of these guarantees undermines the ability of non-technical stakeholders to integrate them into practical systems. Hence, researchers should spend effort to translate their technical metrics into plain-language standards that practitioners can adopt.

    \item \textbf{Impossibility Results}: Theorists can contribute to the study of interactions by establishing formal impossibility results that characterize the conditions under which multiple TAI aspects are genuinely irreconcilable. This will inform users of when trade-off management is necessary, as opposed to pursuing joint optimization.

    \item \textbf{Further Documentation of Interactions}: While this paper surveyed interactions across 5 aspects and selected solutions, there are many solutions and aspects we have not endeavored to cover (\Cref{sec:otheraspects}). There is room for researchers to more comprehensively survey interactions that have already been documented, creating reference material, perhaps in the form of a searchable registry of interactions. Given how recently many interactions were first documented, it is likely that many more interactions remain unidentified. Researchers should continue to probe methods that claim to improve some aspect of trust and disclose interactions, along with proposing mitigations.
\end{enumerate}

Additionally, we offer one research program with potential to advance the goal of improving multiple aspects of trust at once, while navigating negative interactions. Multi-objective optimization~\citep{desideri2012moo, sener2018moo} seeks to find a set of Pareto-optimal solutions to a system that has several competing objectives, and could be applied to improve TAI holistically. Given that each relevant TAI aspect can be quantified with a meaningful metric of progress, we can view each aspect as a dimension in $\mathbb{R}^N$, including one for utility. The na\"ive approach to trust applies a technical solution tailored to one aspect at a time. This can be visualized as moving in the positive direction along one aspect, but behaviour along other axes is uncontrolled. Instead, applying multi-objective optimization techniques to all the parameters of our technical solutions at once would find directions that push out the Pareto frontier. We leave this program for future work.

\subsection{Call to Action: Steps to Improve Trust for Practitioners}
Despite the need for a different approach to TAI research, practitioners are deploying AI today and need to ensure its trustworthiness to the greatest extent possible. Here we provide guidance on how to achieve trust in practice, acknowledging the likelihood of negative interactions, trade-offs, and challenges from combining many objectives:
\begin{enumerate}[leftmargin=*, itemsep=2pt, topsep=2pt]
    \item Prior to model development, enumerate all relevant TAI aspects and prioritize them by importance in the application at hand. Involve stakeholders including model developers, users, and subjects of the model. With so many desirable objectives, it may not be possible to achieve them all simultaneously to the same extent. However, some aspects are likely to be more critical than others in any given context. Hence, deliberate relevancy ranking will help to establish clear priorities as to which aspects cannot be compromised on.

    \item Establish clear metrics and develop automated tests for each relevant aspect when possible, but also recognize soft goals and constraints within the deployment context. Progress towards a goal cannot be measured without clear metrics, and deciding on the appropriate metrics before development will clarify these goals from the outset. Not only can metrics indicate improvements, they can also indicate when issues begin to arise, for example from negative interactions. Automated testing removes selection bias which helps ensure negative effects are caught.

    \item Deliberately analyze how TAI aspects could interact, positively and negatively, before implementing technical solutions or optimizing for metrics. Despite one’s best intentions, applying a new technique to improve some aspect of trust can erode progress in other areas. Many such interactions have been documented and hence can be identified ahead of time  (e.g. see \Cref{sec:pairwise}).

    \item Evaluate the potential risks of negative interactions by quantifying their likelihood and severity. Not all negative interactions are necessarily an issue. If the measured impacts are small, the benefits to some aspects may outweigh the negatives to others. By taking a prudent and quantitative risk management approach, the most probable and severe issues should be given the most attention.

    \item When applying technical solutions to improve any single aspect, perform ablations to measure impact on all other aspects, not just accuracy. Failing to notice when a negative interaction has occurred is the most likely way to unintentionally cause harm. The use of automated testing on every change and for every aspect helps to identify such relapses.

    \item When negative interactions or trade-offs are observed, assess the impacts to each aspect and manage compromises according to the pre-established priorities. Again, it may not be possible in all contexts to improve all relevant aspects of trust simultaneously, leading to the need for compromise. In the face of negative interactions, referring back to the pre-established priorities, estimation of risk severity, and quantitative metrics will help to unblock AI deployments and understand the limitations of a system.
\end{enumerate}

These steps will help to develop a risk-based prioritization of TAI aspects and balance competing constraints, enabling users to proactively anticipate, measure, and mitigate negative interactions. Still, achieving multi-faceted trust is not a solved problem, and we do not have a complete solution. We call for TAI researchers to broaden the scope of their work beyond one or two aspects in isolation, and work towards a holistic model of trust combining many aspects while accounting for interactions.

\subsection{Beyond Pairwise Interactions}
In \Cref{sec:pairwise} we surveyed a wide range of literature that has considered pairwise interactions. We briefly mention work that has gone beyond pairwise interactions by studying the intersection of multiple TAI aspects in the direction we are advocating. \citet{ferry2023sok} systematically review three aspects, fairness, privacy, and explainability, pointing out the isolated nature of prior research, and stop short of considering novel challenges of integrating all three aspects at once. \citet{sharma2020certifai} integrate fairness and robustness into their explainability method by design in the spirit of integrated TAI. Meanwhile, \citet{li2023tai} discuss many TAI aspects including fairness, privacy, robustness, and explainability, building a framework of when to consider each aspect throughout the model lifecycle. While advocating for the combination of many TAI aspects, they do not give detailed insights on the negative interactions that can occur.

\section{Alternative Views}
\label{sec:alternative}

While we take the position that Trustworthy AI must account for interactions, there are alternative positions that could be considered. Primary among them is that TAI improvements may detract from the utility of AI systems. Utility is ultimately the most important aspect of a model---no amount of debiasing or robustness will make a model with poor accuracy useful. It is well known that many of the technical solutions we discussed in \Cref{sec:apsects} to improve TAI aspects have a detrimental effect on model accuracy. This contrary position may claim that improving \emph{any} TAI aspect is not worthwhile if it comes at the cost of diminished accuracy, let alone improving \emph{several} simultaneously with compounded effects. However, a model's utility remains theoretical until it is deployed, which requires adherence to laws and regulations on fairness, data privacy, and more. Without holistic TAI, even the most accurate models may never reach real-world application.

\looseness=-1 It could also be argued that the trade-offs between some TAI aspects are fundamental, and therefore trying to simultaneously improve them is impossible. For instance, we mentioned in \Cref{sec:FP} that fairness and privacy seem to be intrinsically in tension---improving or evaluating fairness typically requires collecting and using private information. Acknowledging these fundamental tensions, we advocated for researchers to identify and characterize them, while practitioners should deliberately prioritize aspects through interdisciplinary consultation of stakeholders, managing interaction risks according to their potential severity.

\looseness=-1 Finally, one might argue that integrating many TAI aspects at once is simply infeasible and unrealistic. Given the unpredictability of negative interactions and challenges implicit in optimizing for many objectives simultaneously, holistic TAI may seem like too daunting of a task. However, AI's increasing importance throughout society behooves us to synthesize the many societal values represented by TAI. The practical guidance we offered in \Cref{sec:position} constitutes a pragmatic approach towards breaking down the challenges of TAI, not by optimizing all objectives simultaneously, but by prioritizing them, evaluating risks, and monitoring for unexpected interactions.

\clearpage 

\bibliographystyle{plainnat}
\bibliography{bib}

@InProceedings{huang2024saps,
  title = 	 {Conformal Prediction for Deep Classifier via Label Ranking},
  author =       {Huang, Jianguo and Xi, Huajun and Zhang, Linjun and Yao, Huaxiu and Qiu, Yue and Wei, Hongxin},
  booktitle = 	 {Proceedings of the 41st International Conference on Machine Learning},
  year = 	 {2024},
}

@InProceedings{guo2017calibration,
  title={On calibration of modern neural networks},
  author={Guo, Chuan and Pleiss, Geoff and Sun, Yu and Weinberger, Kilian Q},
  booktitle={International Conference on Machine Learning},
  pages={1321--1330},
  year={2017},
  organization={PMLR}
}

@article{minderer2021revisiting,
  title={Revisiting the calibration of modern neural networks},
  author={Minderer, Matthias and Djolonga, Josip and Romijnders, Rob and Hubis, Frances and Zhai, Xiaohua and Houlsby, Neil and Tran, Dustin and Lucic, Mario},
  journal={Advances in Neural Information Processing Systems},
  volume={34},
  pages={15682--15694},
  year={2021}
}

@article{angelopoulos2022gentle,
  title={A gentle introduction to conformal prediction and distribution-free uncertainty quantification},
  author={Angelopoulos, Anastasios N and Bates, Stephen},
  journal={arXiv:2107.07511},
  year={2021}
}

@InProceedings{cresswell2024icml,
  title = 	 {{Conformal Prediction Sets Improve Human Decision Making}},
  author =       {Cresswell, Jesse C. and Sui, Yi and Kumar, Bhargava and Vouitsis, Noël},
  booktitle = 	 {Proceedings of the 41st International Conference on Machine Learning},
  year = 	 {2024},
}

@inproceedings{
cresswell2024conformal,
title={{Conformal Prediction Sets Can Cause Disparate Impact}},
author={Cresswell, Jesse C. and Kumar, Bhargava and Sui, Yi and Belbahri, Mouloud},
booktitle={International Conference on Learning Representations},
year={2025},
}

@book{vovk2005algorithmic,
  title={Algorithmic Learning in a Random World},
  author={Vovk, Vladimir and Gammerman, Alexander and Shafer, Glenn},
  year={2005},
  publisher={Springer}
}

@book{soize2017uncertainty,
  title={Uncertainty Quantification},
  author={Soize, Christian},
  year={2017},
  publisher={Springer}
}

@article{boenisch2021gradient,
  title={Gradient masking and the underestimated robustness threats of differential privacy in deep learning},
  author={Boenisch, Franziska and Sperl, Philip and B{\"o}ttinger, Konstantin},
  journal={arXiv:2105.07985},
  year={2021}
}

@inproceedings{Abadi_2016,
	doi = {10.1145/2976749.2978318},
	year = 2016,
	publisher = {{ACM}},
	author = {Martin Abadi and Andy Chu and Ian Goodfellow and H. Brendan McMahan and Ilya Mironov and Kunal Talwar and Li Zhang},
	title = {Deep Learning with Differential Privacy},
	booktitle = {Proceedings of the 2016 {ACM} {SIGSAC} Conference on Computer and Communications Security}
}

@inproceedings{
angelopoulos2021raps,
title={Uncertainty Sets for Image Classifiers using Conformal Prediction},
author={Anastasios Nikolas Angelopoulos and Stephen Bates and Michael Jordan and Jitendra Malik},
booktitle={International Conference on Learning Representations},
year={2021},
}

@InProceedings{zhang2019interp,
  title = 	 {Interpreting Adversarially Trained Convolutional Neural Networks},
  author =       {Zhang, Tianyuan and Zhu, Zhanxing},
  booktitle = 	 {Proceedings of the 36th International Conference on Machine Learning},
  year = 	 {2019},
  volume = 	 {97},
}

@INPROCEEDINGS{Carlini2018audio,
  author={Carlini, Nicholas and Wagner, David},
  booktitle={2018 IEEE Security and Privacy Workshops}, 
  title={Audio Adversarial Examples: Targeted Attacks on Speech-to-Text}, 
  year={2018},
  pages={1-7},
  doi={10.1109/SPW.2018.00009}
}

@inproceedings{jia2017text,
    title = "Adversarial Examples for Evaluating Reading Comprehension Systems",
    author = "Jia, Robin  and
      Liang, Percy",
    booktitle = "Proceedings of the 2017 Conference on Empirical Methods in Natural Language Processing",
    year = "2017",
    doi = "10.18653/v1/D17-1215",
}

@inproceedings{ilyas2019bugsnotfeatures,
 author = {Ilyas, Andrew and Santurkar, Shibani and Tsipras, Dimitris and Engstrom, Logan and Tran, Brandon and Madry, Aleksander},
 booktitle = {Advances in Neural Information Processing Systems},
 title = {Adversarial Examples Are Not Bugs, They Are Features},
 volume = {32},
 year = {2019}
}

@article{
xi2024does,
title={{Does confidence calibration improve conformal prediction?}},
author={Huajun Xi and Jianguo Huang and Kangdao Liu and Lei Feng and Hongxin Wei},
journal={Transactions on Machine Learning Research},
issn={2835-8856},
year={2025},
}

@article{huang2024explaining,
  title={Explaining the model, protecting your data: Revealing and mitigating the data privacy risks of post-hoc model explanations via membership inference},
  author={Huang, Catherine and Pawelczyk, Martin and Lakkaraju, Himabindu},
  journal={arXiv:2407.17663},
  year={2024}
}

@article{quan2022amplification,
  title={On the amplification of security and privacy risks by post-hoc explanations in machine learning models},
  author={Quan, Pengrui and Chakraborty, Supriyo and Jeyakumar, Jeya Vikranth and Srivastava, Mani},
  journal={arXiv:2206.14004},
  year={2022}
}

@inproceedings{lakkaraju2020explain,
author = {Lakkaraju, Himabindu and Bastani, Osbert},
title = {{``How do I fool you?'': Manipulating User Trust via Misleading Black Box Explanations}},
year = {2020},
isbn = {9781450371100},
doi = {10.1145/3375627.3375833},
booktitle = {Proceedings of the AAAI/ACM Conference on AI, Ethics, and Society},
pages = {79–85},
numpages = {7},
}

@inproceedings{slack2020robustexplain,
author = {Slack, Dylan and Hilgard, Sophie and Jia, Emily and Singh, Sameer and Lakkaraju, Himabindu},
title = {{Fooling LIME and SHAP: Adversarial Attacks on Post hoc Explanation Methods}},
year = {2020},
isbn = {9781450371100},
doi = {10.1145/3375627.3375830},
booktitle = {Proceedings of the AAAI/ACM Conference on AI, Ethics, and Society},
pages = {180–186},
numpages = {7},
}

@InProceedings{dwork2006,
author="Dwork, Cynthia
and McSherry, Frank
and Nissim, Kobbi
and Smith, Adam",
title="{Calibrating Noise to Sensitivity in Private Data Analysis}",
booktitle="Theory of Cryptography",
year="2006",
publisher="Springer Berlin Heidelberg",
pages="265--284",
isbn="978-3-540-32732-5"
}

@article{liu2021privacy,
author = {Liu, Bo and Ding, Ming and Shaham, Sina and Rahayu, Wenny and Farokhi, Farhad and Lin, Zihuai},
title = {{When Machine Learning Meets Privacy: A Survey and Outlook}},
year = {2021},
publisher = {Association for Computing Machinery},
volume = {54},
number = {2},
issn = {0360-0300},
doi = {10.1145/3436755},
journal = {ACM Comput. Surv.},
articleno = {31},
numpages = {36},
}

@ARTICLE{ye2022privacy,
  author={Ye, Dayong and Shen, Sheng and Zhu, Tianqing and Liu, Bo and Zhou, Wanlei},
  journal={IEEE Transactions on Information Forensics and Security}, 
  title={One Parameter Defense—Defending Against Data Inference Attacks via Differential Privacy}, 
  year={2022},
  volume={17},
  pages={1466-1480},
  doi={10.1109/TIFS.2022.3163591}
}

@inproceedings{biggio2013evasion,
  title={Evasion attacks against machine learning at test time},
  author={Biggio, Battista and Corona, Igino and Maiorca, Davide and Nelson, Blaine and {\v{S}}rndi{\'c}, Nedim and Laskov, Pavel and Giacinto, Giorgio and Roli, Fabio},
  booktitle={Machine Learning and Knowledge Discovery in Databases: European Conference},
  pages={387--402},
  year={2013},
}

@inproceedings{shokri2021privacyexplain,
author = {Shokri, Reza and Strobel, Martin and Zick, Yair},
title = {On the Privacy Risks of Model Explanations},
year = {2021},
isbn = {9781450384735},
doi = {10.1145/3461702.3462533},
booktitle = {Proceedings of the 2021 AAAI/ACM Conference on AI, Ethics, and Society},
pages = {231–241},
numpages = {11},
}

@INPROCEEDINGS{long2020privatefair,
  author={Long, Yunhui and Wang, Lei and Bu, Diyue and Bindschaedler, Vincent and Wang, Xiaofeng and Tang, Haixu and Gunter, Carl A. and Chen, Kai},
  booktitle={2020 IEEE European Symposium on Security and Privacy}, 
  title={A Pragmatic Approach to Membership Inferences on Machine Learning Models}, 
  year={2020},
  pages={521-534},
  doi={10.1109/EuroSP48549.2020.00040}
}

@InProceedings{benz2021robustfair,
  title = 	 {Robustness May Be at Odds with Fairness: An Empirical Study on Class-wise Accuracy},
  author =       {Benz, Philipp and Zhang, Chaoning and Karjauv, Adil and Kweon, In So},
  booktitle = 	 {NeurIPS 2020 Workshop on Pre-registration in Machine Learning},
  pages = 	 {325--342},
  year = 	 {2021},
  volume = 	 {148},
}

@INPROCEEDINGS{chang2021privatefair,
  author={Chang, Hongyan and Shokri, Reza},
  booktitle={2021 IEEE European Symposium on Security and Privacy}, 
  title={On the Privacy Risks of Algorithmic Fairness}, 
  year={2021},
  pages={292-303},
  doi={10.1109/EuroSP51992.2021.00028}}

@inproceedings{dai2022fairexplain,
author = {Dai, Jessica and Upadhyay, Sohini and Aivodji, Ulrich and Bach, Stephen H. and Lakkaraju, Himabindu},
title = {Fairness via Explanation Quality: Evaluating Disparities in the Quality of Post hoc Explanations},
year = {2022},
isbn = {9781450392471},
doi = {10.1145/3514094.3534159},
booktitle = {Proceedings of the 2022 AAAI/ACM Conference on AI, Ethics, and Society},
pages = {203–214},
}

@inproceedings{balagopalan2022explainfair,
author = {Balagopalan, Aparna and Zhang, Haoran and Hamidieh, Kimia and Hartvigsen, Thomas and Rudzicz, Frank and Ghassemi, Marzyeh},
title = {The Road to Explainability is Paved with Bias: Measuring the Fairness of Explanations},
year = {2022},
isbn = {9781450393522},
publisher = {Association for Computing Machinery},
doi = {10.1145/3531146.3533179},
booktitle = {Proceedings of the 2022 ACM Conference on Fairness, Accountability, and Transparency},
pages = {1194–1206},
}

@article{zhu2024uncertainty,
  author={Zhu, Meiyi and Guo, Caili and Feng, Chunyan and Simeone, Osvaldo},
  journal={IEEE Transactions on Information Forensics and Security}, 
  title={{On the Impact of Uncertainty and Calibration on Likelihood-Ratio Membership Inference Attacks}}, 
  year={2025},
  volume={20},
  doi={10.1109/TIFS.2025.3578931}}

@article{ferry2023sok,
author = {Ferry, Julien and Aïvodji, Ulrich and Gambs, Sébastien and Huguet, Marie-José and Siala, Mohamed},
title = {{Taming the Triangle: On the Interplays Between Fairness, Interpretability, and Privacy in Machine Learning}},
journal = {Computational Intelligence},
volume = {41},
number = {4},
doi = {https://doi.org/10.1111/coin.70113},
year = {2025}
}

@article{
bu2023on,
title={On the Convergence and Calibration of Deep Learning with Differential Privacy},
author={Zhiqi Bu and Hua Wang and Zongyu Dai and Qi Long},
journal={Transactions on Machine Learning Research},
issn={2835-8856},
year={2023},
}

@article{zhang2022closer,
  title={A closer look at the calibration of differentially private learners},
  author={Zhang, Hanlin and Li, Xuechen and Sen, Prithviraj and Roukos, Salim and Hashimoto, Tatsunori},
  journal={arXiv:2210.08248},
  year={2022}
}

@article{Angelopoulos2022Private,
	author = {Angelopoulos, Anastasios Nikolas and Bates, Stephen and Zrnic, Tijana and Jordan, Michael I.},
	journal = {Harvard Data Science Review},
	number = {2},
	year = {2022},
	publisher = {The MIT Press},
	title = {Private {Prediction} {Sets}},
	volume = {4},
}

@inproceedings{ICLR2024_83b7da3e,
 author = {Qi, Xiangyu and Zeng, Yi and Xie, Tinghao and Chen, Pin-Yu and Jia, Ruoxi and Mittal, Prateek and Henderson, Peter},
 booktitle = {International Conference on Learning Representations},
 title = {{Fine-tuning Aligned Language Models Compromises Safety, Even When Users Do Not Intend To!}},
 volume = {2024},
 year = {2024}
}

@article{desideri2012moo,
     author = {Jean-Antoine D\'esid\'eri},
     title = {{Multiple-gradient descent algorithm (MGDA) for multiobjective optimization}},
     journal = {Comptes Rendus. Math\'ematique},
     year = {2012},
     publisher = {Elsevier},
     volume = {350},
     number = {5-6},
     doi = {10.1016/j.crma.2012.03.014},
}

@inproceedings{sener2018moo,
 author = {Sener, Ozan and Koltun, Vladlen},
 booktitle = {Advances in Neural Information Processing Systems},
 title = {{Multi-Task Learning as Multi-Objective Optimization}},
 volume = {31},
 year = {2018}
}

@article{mittelstadt2024unfairness,
  author  = {Mittelstadt, Brent and Wachter, Sandra and Russell, Chris},
  title   = {{The Unfairness of Fair Machine Learning: Leveling Down and Strict Egalitarianism by Default}},
  journal = {Michigan Technology Law Review},
  volume  = {30},
  number  = {1},
  year    = {2024},
  doi     = {10.36645/mtlr.30.1.unfairness},
}

@inproceedings{mahmoud2026unintended,
    title = "{The Unintended Trade-off of AI Alignment: Balancing Hallucination Mitigation and Safety in LLMs}",
    author = "Mahmoud, Omar  and
      Khalil, Ali  and
      Karimpanal, Thommen George  and
      Semage, Buddhika Laknath  and
      Rana, Santu",
    booktitle = "Findings of the {A}ssociation for {C}omputational {L}inguistics: {EACL} 2026",
    year = "2026",
    publisher = "Association for Computational Linguistics",
    doi = "10.18653/v1/2026.findings-eacl.53",
    ISBN = "979-8-89176-386-9",
}

@inproceedings{
szegedy2014intriguing,
  title={Intriguing properties of neural networks},
  author={Szegedy, Christian and Zaremba, Wojciech and Sutskever, Ilya and Bruna, Joan and Erhan, Dumitru and Goodfellow, Ian and Fergus, Rob},
booktitle={International Conference on Learning Representations},
year={2014},
}

@inproceedings{nanda2021robustfair,
author = {Nanda, Vedant and Dooley, Samuel and Singla, Sahil and Feizi, Soheil and Dickerson, John P.},
title = {Fairness Through Robustness: Investigating Robustness Disparity in Deep Learning},
year = {2021},
isbn = {9781450383097},
booktitle = {Proceedings of the 2021 ACM Conference on Fairness, Accountability, and Transparency},
pages = {466–477},
numpages = {12},
}

@InProceedings{xu2021robustfair,
  title = 	 {To be Robust or to be Fair: Towards Fairness in Adversarial Training},
  author =       {Xu, Han and Liu, Xiaorui and Li, Yaxin and Jain, Anil and Tang, Jiliang},
  booktitle = 	 {Proceedings of the 38th International Conference on Machine Learning},
  year = 	 {2021},
  volume = 	 {139},
  series = 	 {Proceedings of Machine Learning Research},
  publisher =    {PMLR},
}

@article{chang2020adversarial,
  title={On adversarial bias and the robustness of fair machine learning},
  author={Chang, Hongyan and Nguyen, Ta Duy and Murakonda, Sasi Kumar and Kazemi, Ehsan and Shokri, Reza},
  journal={arXiv:2006.08669},
  year={2020}
}

@article{saifullah2024privacy,
  title={The privacy-explainability trade-off: Unraveling the impacts of differential privacy and federated learning on attribution methods},
  author={Saifullah, Saifullah and Mercier, Dominique and Lucieri, Adriano and Dengel, Andreas and Ahmed, Sheraz},
  journal={Frontiers in Artificial Intelligence},
  volume={7},
  pages={1236947},
  year={2024},
  publisher={Frontiers Media SA}
}

@inproceedings{patel2022dpxai,
author = {Patel, Neel and Shokri, Reza and Zick, Yair},
title = {Model Explanations with Differential Privacy},
year = {2022},
isbn = {9781450393522},
publisher = {Association for Computing Machinery},
doi = {10.1145/3531146.3533235},
booktitle = {Proceedings of the 2022 ACM Conference on Fairness, Accountability, and Transparency},
pages = {1895–1904},
numpages = {10},
}

@Article{linardatos2021ex,
AUTHOR = {Linardatos, Pantelis and Papastefanopoulos, Vasilis and Kotsiantis, Sotiris},
TITLE = {{Explainable AI: A Review of Machine Learning Interpretability Methods}},
JOURNAL = {Entropy},
VOLUME = {23},
YEAR = {2021},
NUMBER = {1},
ARTICLE-NUMBER = {18},
PubMedID = {33375658},
ISSN = {1099-4300},
DOI = {10.3390/e23010018}
}

@inproceedings{sharma2020certifai,
author = {Sharma, Shubham and Henderson, Jette and Ghosh, Joydeep},
title = {{CERTIFAI: A Common Framework to Provide Explanations and Analyse the Fairness and Robustness of Black-box Models}},
year = {2020},
isbn = {9781450371100},
doi = {10.1145/3375627.3375812},
booktitle = {Proceedings of the AAAI/ACM Conference on AI, Ethics, and Society},
pages = {166–172},
numpages = {7},
}

@inproceedings{
madry2018towards,
title={Towards Deep Learning Models Resistant to Adversarial Attacks},
author={Aleksander Madry and Aleksandar Makelov and Ludwig Schmidt and Dimitris Tsipras and Adrian Vladu},
booktitle={International Conference on Learning Representations},
year={2018},
}

@article{islam2021explainable,
  title={{Explainable artificial intelligence approaches: A survey}},
  author={Islam, Sheikh Rabiul and Eberle, William and Ghafoor, Sheikh Khaled and Ahmed, Mohiuddin},
  journal={arXiv:2101.09429},
  year={2021}
}

@inproceedings{
antoran2021getting,
title={{Getting a CLUE: A  Method for Explaining Uncertainty Estimates}},
author={Javier Antoran and Umang Bhatt and Tameem Adel and Adrian Weller and Jos{\'e} Miguel Hern{\'a}ndez-Lobato},
booktitle={International Conference on Learning Representations},
year={2021},
}

@article{ZHOU2025109681,
title = {Advancing explainability of adversarial trained Convolutional Neural Networks for robust engineering applications},
journal = {Engineering Applications of Artificial Intelligence},
volume = {140},
pages = {109681},
year = {2025},
issn = {0952-1976},
doi = {10.1016/j.engappai.2024.109681},
author = {Dehua Zhou and Ziyu Song and Zicong Chen and Xianting Huang and Congming Ji and Saru Kumari and Chien-Ming Chen and Sachin Kumar},
}

@article{rahman2018membership,
  title={Membership inference attack against differentially private deep learning model},
  author={Rahman, Md Atiqur and Rahman, Tanzila and Lagani{\`e}re, Robert and Mohammed, Noman and Wang, Yang},
journal = {Transactions on Data Privacy},
volume={11},
number={1},
year={2018}
}

@inproceedings{ribeiro2016lime,
author = {Ribeiro, Marco Tulio and Singh, Sameer and Guestrin, Carlos},
title = {{"Why Should I Trust You?": Explaining the Predictions of Any Classifier}},
year = {2016},
isbn = {9781450342322},
doi = {10.1145/2939672.2939778},
booktitle = {Proceedings of the 22nd ACM SIGKDD International Conference on Knowledge Discovery and Data Mining},
pages = {1135–1144},
numpages = {10},
}

@inproceedings{
gendler2022adversarially,
title={Adversarially Robust Conformal Prediction},
author={Asaf Gendler and Tsui-Wei Weng and Luca Daniel and Yaniv Romano},
booktitle={International Conference on Learning Representations},
year={2022},
}

@article{romano2020with,
	author = {Romano, Yaniv and Barber, Rina Foygel and Sabatti, Chiara and Cand{\` e}s, Emmanuel},
	journal = {Harvard Data Science Review},
	number = {2},
	year = {2020},
	title = {With {Malice} {Toward} {None}: Assessing {Uncertainty} via {Equalized} {Coverage}},
	volume = {2},
}

@inproceedings{ye2022mia,
author = {Ye, Jiayuan and Maddi, Aadyaa and Murakonda, Sasi Kumar and Bindschaedler, Vincent and Shokri, Reza},
title = {Enhanced Membership Inference Attacks against Machine Learning Models},
year = {2022},
isbn = {9781450394505},
doi = {10.1145/3548606.3560675},
booktitle = {Proceedings of the 2022 ACM SIGSAC Conference on Computer and Communications Security},
pages = {3093–3106},
}

@misc{occ_fair_lending,
  author       =  {{Office of the Comptroller of the Currency}},
year = {2025},
title = {Fair Lending},
url = {https://www.occ.treas.gov/topics/consumers-and-communities/consumer-protection/fair-lending/index-fair-lending.html}, note = {Accessed: 2025-01-01} }

@article{gibbs2023conformal,
    author = {Gibbs, Isaac and Cherian, John J and Cand{\`e}s, Emmanuel J},
    title = {Conformal prediction with conditional guarantees},
    journal = {Journal of the Royal Statistical Society Series B: Statistical Methodology},
    year = {2025},
    issn = {1369-7412},
    doi = {10.1093/jrsssb/qkaf008},
}

@inproceedings{romano2020classification,
 author = {Romano, Yaniv and Sesia, Matteo and Candes, Emmanuel},
 booktitle = {Advances in Neural Information Processing Systems},
 title = {Classification with Valid and Adaptive Coverage},
 volume = {33},
 year = {2020}
}

@article{ding2024class,
  title={Class-conditional conformal prediction with many classes},
  author={Ding, Tiffany and Angelopoulos, Anastasios and Bates, Stephen and Jordan, Michael and Tibshirani, Ryan J},
  journal={Advances in Neural Information Processing Systems},
  volume={36},
  year={2024}
}

@article{gosiewska2019not,
  title={Do not trust additive explanations},
  author={Gosiewska, Alicja and Biecek, Przemyslaw},
  journal={arXiv:1903.11420},
  year={2019}
}

@article{carey2023reidentification,
author = {Carey, CJ and Dick, Travis and Epasto, Alessandro and Javanmard, Adel and Karlin, Josh and Kumar, Shankar and Mu\~{n}oz Medina, Andres and Mirrokni, Vahab and Nunes, Gabriel Henrique and Vassilvitskii, Sergei and Zhong, Peilin},
title = {Measuring Re-identification Risk},
year = {2023},
publisher = {Association for Computing Machinery},
volume = {1},
number = {2},
doi = {10.1145/3589294},
journal = {Proc. ACM Manag. Data},
articleno = {149},
numpages = {26},
}

@article{zhang2019should,
      title={{"Why Should You Trust My Explanation?" Understanding Uncertainty in LIME Explanations}},  author={Zhang, Yujia and Song, Kuangyan and Sun, Yiming and Tan, Sarah and Udell, Madeleine},
  journal={arXiv:1904.12991},
  year={2019}
}

@InProceedings{ghosh2023robustconformal,
  title = 	 {Probabilistically robust conformal prediction},
  author =       {Ghosh, Subhankar and Shi, Yuanjie and Belkhouja, Taha and Yan, Yan and Doppa, Jana and Jones, Brian},
  booktitle = 	 {Proceedings of the Thirty-Ninth Conference on Uncertainty in Artificial Intelligence},
  pages = 	 {681--690},
  year = 	 {2023},
  volume = 	 {216},
}

@inproceedings{slack2021explainuq,
 author = {Slack, Dylan and Hilgard, Anna and Singh, Sameer and Lakkaraju, Himabindu},
 booktitle = {Advances in Neural Information Processing Systems},
 pages = {9391--9404},
 title = {Reliable Post hoc Explanations: Modeling Uncertainty in Explainability},
 volume = {34},
 year = {2021}
}

@article{lofstrom2024explaincalibration,
title = {Calibrated explanations: With uncertainty information and counterfactuals},
journal = {Expert Systems with Applications},
volume = {246},
pages = {123154},
year = {2024},
issn = {0957-4174},
doi = {10.1016/j.eswa.2024.123154},
author = {Helena Löfström and Tuwe Löfström and Ulf Johansson and Cecilia Sönströd},
}

@Inbook{Kindermans2019,
author="Kindermans, Pieter-Jan
and Hooker, Sara
and Adebayo, Julius
and Alber, Maximilian
and Sch{\"u}tt, Kristof T.
and D{\"a}hne, Sven
and Erhan, Dumitru
and Kim, Been",
title="The (Un)reliability of Saliency Methods",
bookTitle="Explainable AI: Interpreting, Explaining and Visualizing Deep Learning",
year="2019",
publisher="Springer International Publishing",
pages="267--280",
isbn="978-3-030-28954-6",
doi="10.1007/978-3-030-28954-6_14",
}

@InProceedings{yapicioglu2024euq,
author="Yapicioglu, Fatima Rabia
and Stramiglio, Alessandra
and Vitali, Fabio",
title="{ConformaSight: Conformal Prediction-Based Global and Model-Agnostic Explainability Framework}",
booktitle="Explainable Artificial Intelligence",
year="2024",
pages="270--293",
isbn="978-3-031-63800-8"
}

@article{bykov2020much,
  title={{How Much Can I Trust You?--Quantifying Uncertainties in Explaining Neural Networks}},
  author={Bykov, Kirill and H{\"o}hne, Marina M-C and M{\"u}ller, Klaus-Robert and Nakajima, Shinichi and Kloft, Marius},
  journal={arXiv:2006.09000},
  year={2020}
}

@article{ahn2023uncertainty,
  title={Uncertainty Quantification for Local Model Explanations Without Model Access},
  author={Ahn, Surin and Grana, Justin and Tamene, Yafet and Holsheimer, Kristian},
  journal={arXiv:2301.05761},
  year={2023}
}

@InProceedings{liu2024conformaladversarial,
  title = 	 {The Pitfalls and Promise of Conformal Inference Under Adversarial Attacks},
  author =       {Liu, Ziquan and Cui, Yufei and Yan, Yan and Xu, Yi and Ji, Xiangyang and Liu, Xue and Chan, Antoni B.},
  booktitle = 	 {Proceedings of the 41st International Conference on Machine Learning},
  pages = 	 {30908--30928},
  year = 	 {2024},
  volume = 	 {235},
}

@article{chouldechova2017fair,
  title     = {{Fair Prediction with Disparate Impact: A Study of Bias in Recidivism Prediction Instruments}},
  author    = {Chouldechova, Alexandra},
  journal   = {Big Data},
  volume    = {5},
  number    = {2},
  year      = {2017},
  doi       = {10.1089/big.2016.0047},
}

@inproceedings{
esipova2023disparate,
title={Disparate Impact in Differential Privacy from Gradient Misalignment},
author={Maria S. Esipova and Atiyeh Ashari Ghomi and Yaqiao Luo and Jesse C. Cresswell},
booktitle={The Eleventh International Conference on Learning Representations },
year={2023}
}

@INPROCEEDINGS{shokri2017,
  author={Shokri, Reza and Stronati, Marco and Song, Congzheng and Shmatikov, Vitaly},
  booktitle={2017 IEEE Symposium on Security and Privacy}, 
  title={Membership Inference Attacks Against Machine Learning Models}, 
  year={2017},
  volume={},
  number={},
  pages={3-18},
  doi={10.1109/SP.2017.41}}

@article{goodfellow2015explaining,
  title={Explaining and harnessing adversarial examples},
  author={Goodfellow, Ian J and Shlens, Jonathon and Szegedy, Christian},
  journal={arXiv:1412.6572},
  year={2014}
}

@inproceedings{tramer2021differentially,
  title={Differentially Private Learning Needs Better Features (or Much More Data)},
  author={Tram{\`e}r, Florian and Boneh, Dan},
  booktitle={International Conference on Learning Representations},
  year={2021}
}

@inproceedings{
tsipras2018robustness,
title={Robustness May Be at Odds with Accuracy},
author={Dimitris Tsipras and Shibani Santurkar and Logan Engstrom and Alexander Turner and Aleksander Madry},
booktitle={International Conference on Learning Representations},
year={2019}
}

@article{yousefpour2021opacus,
  title={{Opacus: User-friendly differential privacy library in PyTorch}},
  author={Yousefpour, Ashkan and Shilov, Igor and Sablayrolles, Alexandre and Testuggine, Davide and Prasad, Karthik and Malek, Mani and Nguyen, John and Ghosh, Sayan and Bharadwaj, Akash and Zhao, Jessica and Cormode, Graham and Mironov, Ilya},
  journal={arXiv:2109.12298},
  year={2021}
}

@article{yeom2020overfitting,
  title={Overfitting, robustness, and malicious algorithms: A study of potential causes of privacy risk in machine learning},
  author={Yeom, Samuel and Giacomelli, Irene and Menaged, Alan and Fredrikson, Matt and Jha, Somesh},
  journal={Journal of Computer Security},
  volume={28},
  number={1},
  pages={35--70},
  year={2020},
  publisher={IOS Press}
}

@article{ghorbani2019explainrobust, 
title={Interpretation of Neural Networks Is Fragile}, 
volume={33},
DOI={10.1609/aaai.v33i01.33013681}, 
number={01}, 
journal={Proceedings of the AAAI Conference on Artificial Intelligence}, 
author={Ghorbani, Amirata and Abid, Abubakar and Zou, James},
year={2019}, 
month={Jul.}, 
pages={3681-3688}
}

@article{hayes2022learning,
  title={Learning to be adversarially robust and differentially private},
  author={Hayes, Jamie and Balle, Borja and Kumar, M Pawan},
  journal={arXiv:2201.02265},
  year={2022}
}

@InProceedings{fawzi2018empirical,
author = {Fawzi, Alhussein and Moosavi-Dezfooli, Seyed-Mohsen and Frossard, Pascal and Soatto, Stefano},
title = {Empirical Study of the Topology and Geometry of Deep Networks},
booktitle = {Proceedings of the IEEE Conference on Computer Vision and Pattern Recognition},
month = {June},
year = {2018}
}

@article{mironov2019renyi,
  title={{Rényi Differential Privacy of the Sampled Gaussian Mechanism}},
  author={Mironov, Ilya and Talwar, Kunal and Zhang, Li},
  journal={arXiv:1908.10530},
  year={2019}
}

@inproceedings{bagdasaryan2019,
  title={Differential privacy has disparate impact on model accuracy},
  author={Bagdasaryan, Eugene and Poursaeed, Omid and Shmatikov, Vitaly},
  booktitle={Advances in Neural Information Processing Systems},
  volume={32},
  pages={15479--15488},
  year={2019}
}

@inproceedings{tran2021a,
  title     = {Decision Making with Differential Privacy under a Fairness Lens},
  author    = {Tran, Cuong and Fioretto, Ferdinando and Van Hentenryck, Pascal and Yao, Zhiyan},
  booktitle = {Proceedings of the Thirtieth International Joint Conference on
               Artificial Intelligence},
  year      = {2021},
  doi       = {10.24963/ijcai.2021/78},
}

@inproceedings{tian2024fairprivate,
  title     = {When Fairness Meets Privacy: Exploring Privacy Threats in Fair Binary Classifiers via Membership Inference Attacks},
  author    = {Tian, Huan and Zhang, Guangsheng and Liu, Bo and Zhu, Tianqing and Ding, Ming and Zhou, Wanlei},
  booktitle = {Proceedings of the Thirty-Third International Joint Conference on
               Artificial Intelligence},
  year      = {2024},
  doi       = {10.24963/ijcai.2024/57},
}

@inproceedings{tran2024fairrobust,
  title     = {On the Effects of Fairness to Adversarial Vulnerability},
  author    = {Tran, Cuong and Zhu, Keyu and Van Hentenryck, Pascal and Fioretto, Ferdinando},
  booktitle = {Proceedings of the Thirty-Third International Joint Conference on
               Artificial Intelligence},
  year      = {2024},
  doi       = {10.24963/ijcai.2024/58},
}

@InProceedings{straitouri2024designing,
  title = 	 {Designing Decision Support Systems using Counterfactual Prediction Sets},
  author =       {Straitouri, Eleni and Gomez Rodriguez, Manuel},
  booktitle = 	 {Proceedings of the 41st International Conference on Machine Learning},
  year = 	 {2024},
  volume = 	 {235},
}

@article{vovk2003mondrian,
  title={Mondrian confidence machine},
  author={Vovk, Vladimir and Lindsay, David and Nouretdinov, Ilia and Gammerman, Alex},
  journal={Technical Report},
  year={2003}
}

@article{
wu2024augment,
title={Augment then Smooth: Reconciling Differential Privacy with Certified Robustness},
author={Jiapeng Wu and Atiyeh Ashari Ghomi and David Glukhov and Jesse C. Cresswell and Franziska Boenisch and Nicolas Papernot},
journal={Transactions on Machine Learning Research},
issn={2835-8856},
year={2024},
}

@inproceedings{carlini2022onion,
 author = {Carlini, Nicholas and Jagielski, Matthew and Zhang, Chiyuan and Papernot, Nicolas and Terzis, Andreas and Tramer, Florian},
 booktitle = {Advances in Neural Information Processing Systems},
 pages = {13263--13276},
 title = {The Privacy Onion Effect: Memorization is Relative},
 volume = {35},
 year = {2022}
}

@article{kulynych2022disparate,
  title={Disparate Vulnerability to Membership Inference Attacks},
  author={Kulynych, Bogdan and Yaghini, Mohammad and Cherubin, Giovanni and Veale, Michael and Troncoso, Carmela},
  journal={Proceedings on Privacy Enhancing Technologies Symposium},
  year={2022},
doi = {10.2478/popets-2022-0023},
}

@article{tonni2020data,
  title={Data and model dependencies of membership inference attack},
  author={Tonni, Shakila Mahjabin and Vatsalan, Dinusha and Farokhi, Farhad and Kaafar, Dali and Lu, Zhigang and Tangari, Gioacchino},
  journal={arXiv:2002.06856},
  year={2020}
}

@book{hendrycks2024safety,
  title={Introduction to AI Safety, Ethics and Society},
  author={Hendrycks, Dan},
  year={2024},
  publisher={Taylor \& Francis},
  isbn={9781032798028},
}

@inproceedings{song2019privaterobust,
author = {Song, Liwei and Shokri, Reza and Mittal, Prateek},
title = {Privacy Risks of Securing Machine Learning Models against Adversarial Examples},
year = {2019},
isbn = {9781450367479},
doi = {10.1145/3319535.3354211},
booktitle = {Proceedings of the 2019 ACM SIGSAC Conference on Computer and Communications Security},
pages = {241–257},
numpages = {17},
}

@article{amodei2016concrete,
  title={{Concrete problems in AI safety}},
  author={Amodei, Dario and Olah, Chris and Steinhardt, Jacob and Christiano, Paul and Schulman, John and Man{\'e}, Dan},
  journal={arXiv:1606.06565},
  year={2016}
}

@book{russell2019human,
  title={Human Compatible: AI and the Problem of Control},
  author={Russell, Stuart},
  year={2019},
  publisher={Penguin UK}
}

@article{gabriel2020artificial,
  title={{Artificial Intelligence, Values, and Alignment}},
  author={Gabriel, Iason},
  journal={Minds and Machines},
  volume={30},
  number={3},
  pages={411--437},
  year={2020},
  publisher={Springer}
}

@InProceedings{sorensen2024alignment,
  title = 	 {Position: A Roadmap to Pluralistic Alignment},
  author =       {Sorensen, Taylor and Moore, Jared and Fisher, Jillian and Gordon, Mitchell L and Mireshghallah, Niloofar and Rytting, Christopher Michael and Ye, Andre and Jiang, Liwei and Lu, Ximing and Dziri, Nouha and Althoff, Tim and Choi, Yejin},
  booktitle = 	 {Proceedings of the 41st International Conference on Machine Learning},
  year = 	 {2024},
  volume = 	 {235},
}

@article{pineau2021reproducibility,
  author  = {Joelle Pineau and Philippe Vincent-Lamarre and Koustuv Sinha and Vincent Lariviere and Alina Beygelzimer and Florence d'Alche-Buc and Emily Fox and Hugo Larochelle},
  title   = {{Improving Reproducibility in Machine Learning Research (A Report from the NeurIPS 2019 Reproducibility Program)}},
  journal = {Journal of Machine Learning Research},
  year    = {2021},
  volume  = {22},
  number  = {164},
  pages   = {1--20},
}

@InProceedings{buolamwini2018gender,
  title = 	 {Gender Shades: Intersectional Accuracy Disparities in Commercial Gender Classification},
  author = 	 {Buolamwini, Joy and Gebru, Timnit},
  booktitle = 	 {Proceedings of the 1st Conference on Fairness, Accountability and Transparency},
  pages = 	 {77--91},
  year = 	 {2018},
  volume = 	 {81},
  series = 	 {Proceedings of Machine Learning Research},
  month = 	 {23--24 Feb},
  publisher =    {PMLR},
}

@article{fazelpour2022diversity,
  title={Diversity in sociotechnical machine learning systems},
  author={Fazelpour, Sina and De-Arteaga, Maria},
  journal={Big Data \& Society},
  volume={9},
  number={1},
  pages={20539517221082027},
  year={2022},
  publisher={SAGE Publications Sage UK: London, England}
}

@inproceedings{cooper2022accountability,
author = {Cooper, A. Feder and Moss, Emanuel and Laufer, Benjamin and Nissenbaum, Helen},
title = {Accountability in an Algorithmic Society: Relationality, Responsibility, and Robustness in Machine Learning},
year = {2022},
isbn = {9781450393522},
doi = {10.1145/3531146.3533150},
booktitle = {Proceedings of the 2022 ACM Conference on Fairness, Accountability, and Transparency},
pages = {864–876},
numpages = {13},
}

@article{li2023tai,
author = {Li, Bo and Qi, Peng and Liu, Bo and Di, Shuai and Liu, Jingen and Pei, Jiquan and Yi, Jinfeng and Zhou, Bowen},
title = {{Trustworthy AI: From Principles to Practices}},
year = {2023},
volume = {55},
number = {9},
issn = {0360-0300},
doi = {10.1145/3555803},
journal = {ACM Comput. Surv.},
articleno = {177},
numpages = {46},
}

@article{fanni2023enhancing,
  title={Enhancing human agency through redress in Artificial Intelligence Systems},
  author={Fanni, Rosanna and Steinkogler, Valerie Eveline and Zampedri, Giulia and Pierson, Jo},
  journal={AI \& Society},
  volume={38},
  number={2},
  pages={537--547},
  year={2023},
  publisher={Springer}
}

\end{document}